%% file: main.tex
\documentclass[11pt,a4paper]{article}
\usepackage[hyperref]{eacl2021}
\usepackage{times}
\usepackage{amsmath}
\usepackage{amsfonts}
\usepackage{latexsym}
\usepackage{booktabs}
\usepackage{url}
\usepackage{tikz}
\usepackage{multirow}
\usepackage{qtree}
\usepackage{float}
\usepackage{tikz-cd}
\usepackage{pgfplots}
\usepackage[]{todonotes}
\usepackage[caption=false]{subfig}

\usepackage{enumitem}
\aclfinalcopy

\pgfplotsset{every axis/.append style={
                    label style={font=\footnotesize},
                    tick label style={font=\footnotesize} 
                    }}

\title{Elastic weight consolidation for better bias inoculation}
\author{James Thorne \\
  Department of Computer Science \\
  University of Cambridge \\
  {\tt jt719@cam.ac.uk} \\\And
  Andreas Vlachos \\
  Department of Computer Science \\
  University of Cambridge \\
  {\tt av308@cam.ac.uk} \\}

\date{}

\begin{document}
\maketitle
\begin{abstract}
The biases present in training datasets have been shown to affect models for sentence pair classification tasks such as natural language inference (NLI) and fact verification. 
While fine-tuning models on additional data has been used to mitigate them, a common issue is that of catastrophic forgetting of the original training dataset.
In this paper, we show that elastic weight consolidation (EWC) allows fine-tuning of models to mitigate biases while being less susceptible to catastrophic forgetting. 
In our evaluation on fact verification and NLI stress tests, we show that fine-tuning with EWC
dominates standard fine-tuning, yielding models with lower levels of forgetting on the original (biased) dataset for equivalent gains in accuracy on the fine-tuning (unbiased) dataset. 

%Furthermore, we demonstrate compatibility between 
% task.
%method of instance re-weighting when mitigating hypothesis only bias in the FEVER shared task. \todo{Given that their results are not directly comparable, I think it makes sense to just combine this sentence with the next one, and maybe add one more nice thing about our paper, e.g. the pareto optimality as a way of comparing approaches, or something else }
%Additionally, we show that systems trained on NLI can be fine-tuned to improve their accuracy on stress test challenge tasks with minimal loss in accuracy on the MultiNLI dataset despite greater domain shift.%\todo{The abstract is ok, but I would avoid extreme; maybe say what it caused to previous work?}
%\todo[inline]{Can be combined with debiased fine-tuning for model-based approaches.}
%
\end{abstract}

\section{Introduction}
A number of recent works have illustrated shortcomings in sentence-pair classification models that are used for Natural Language Inference (NLI).
These arise from limited or biased training data and the lack of suitable inductive bias in models.
\citet{Naik2018} demonstrated that phenomena such as the presence of negation or a high degree of lexical overlap induce misclassifications on models trained on the MultiNLI dataset \citep{Williams2018}.
\citet{Poliak2018} and \citet{Gururangan2018} identified biases introduced during dataset construction that were exploited by models to learn associations between the label and only one of the two input sentences without considering the other -- known as \emph{hypothesis-only} bias.

%
%Mitigating these biases is especially pertinent in the task of fact verification: predicting whether a claim is supported or refuted by evidence. 
%It is critical that evidence is correctly taken into account when labeling an instance or generating an explanation to the end-user. A common model for fact verification \citep{FNC,Thorne2018a} is text-pair classification between a claim and evidence retrieved from a trusted source \citep[\em inter alia]{Nie2018, Yoneda2018, Hanselowski2018}. 
%However, like NLI datasets, it has been observed \citep{Schuster2019} that cognitive shortcuts taken by annotators when constructing instances manifest themselves as giveaway cues that models learn to exploit to classify the text-pair with reasonable accuracy by considering only the claim and without considering the evidence.
Such biases also affect fact verification \citep{Schuster2019}, typically modeled as a text-pair classification between a claim and evidence retrieved from a trusted source \citep{FNC,Thorne2018a}.

\begin{figure}[H]
    \centering
    \includegraphics[width=\linewidth]{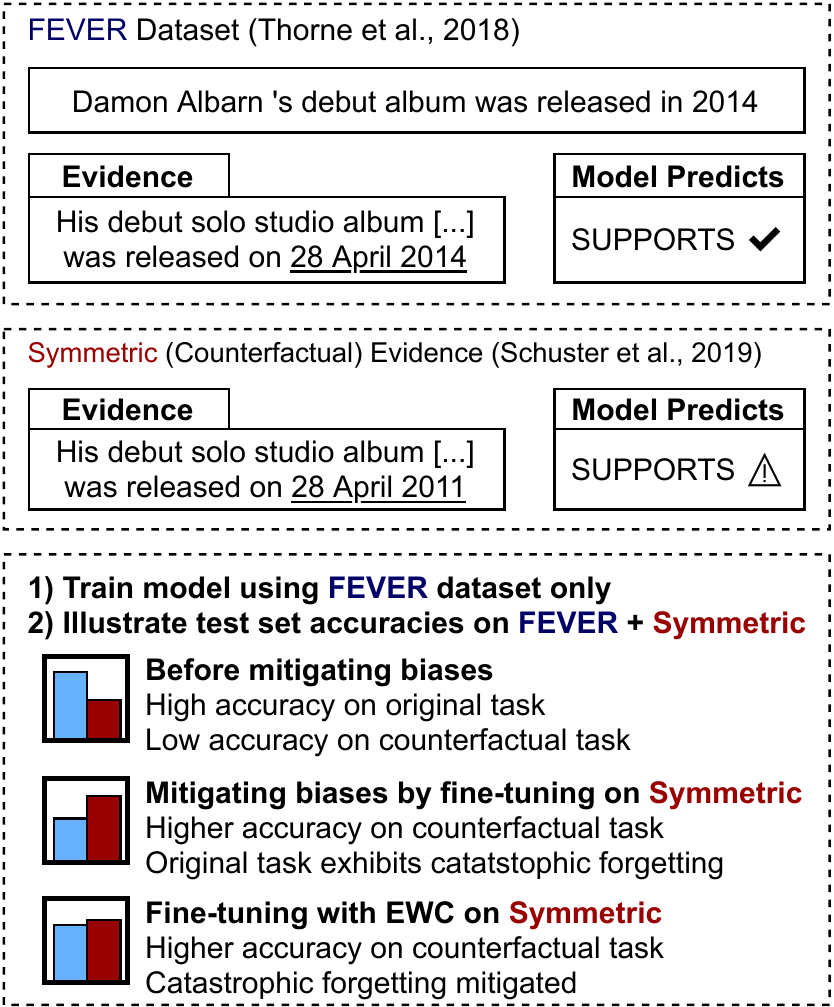}
    \caption{Hypothesis only bias in FEVER contributes to low accuracy when testing against counterfactual evidence. This is mitigated by fine-tuning on counterfactual evidence. Catastrophic forgetting from fine-tuning is reduced when using elastic weight consolidation (EWC), preserving the original task accuracy. }
    \label{fig:my_label}
\end{figure}

To mitigate these undesirable behaviors %(caused by limitations in the training data and lack of suitable inductive bias in the models),
\citet{Liu2019InoculationDatasets} fine-tune models with a small targeted number of labeled instances to ``inoculate'' the models. This can be contrasted to methods such as Debiased Focal Loss (DFL) and Product of Experts (POE) \cite{Mahabadi2020End-to-EndCorpora} which require architectural changes to separately encode and penalize biases. 
In related work, \citet{Suntwal2019} delexicalize instances -- replacing tokens with placeholders, preventing classifiers from exploiting mutual information between domain-specific noun-phrases and class labels.
Finally, \citet{Schuster2019} re-weight the loss of instances during training guided by the mutual information between tokens and the instance labels.
Each of these methods corresponds to a multi-objective optimization problem where the original dataset accuracy is often sacrificed in favor of accuracy on a different evaluation set.%, 

For fine-tuning specifically, the reduction in accuracy on the original task called catastrophic forgetting \citep{French1999} as parameters for the original task are overridden during fine-tuning.
In this paper, we show that regularizing fine-tuning with Elastic Weight Consolidation \citep[EWC]{Kirkpatrick2017} minimizes catastrophic forgetting by penalizing weight updates to parameters crucial to modeling the original dataset. 
EWC has previously been applied to machine translation, where \citet{Thompson2019} and \citet{Saunders2019} minimized catastrophic forgetting for models undergoing domain adaption. %through fine-tuning
Extending this line of research further, we demonstrate that EWC can be used to mitigate biases present in two sentence-pair classification tasks without architectural changes to the models.
We evaluate a number of popular model architectures, trained on MultiNLI and FEVER, and demonstrate that fine-tuning with EWC mitigates model bias while yielding %significantly 
lower reductions in accuracy on the original dataset compared to standard fine-tuning.

%Our experiments\todo{maybe start with a sentence saying that we do two sets of epxeriments and then go into each of them} show that using as few as 300 debiased instances from the Symmetric dataset released by \citet{Schuster2019} is sufficient to mitigate hypothesis-only bias in systems trained on the FEVER dataset through fine-tuning.\todo{Nice but not the main point which is avoiding catastrophic forgetting. Save for later?}
%minimizing impact on the original task.
%On both MultiNLI and FEVER, we show that EWC mitigates catastrophic forgetting 

On all experiments on the FEVER dataset, fine-tuning with symmetric counterfactual data \citep{Schuster2019} mitigated hypothesis-only bias, increasing the absolute accuracy of a BERT model by approximately $10\%$. 
Without EWC, accuracy on the original dataset was reduced from $87\%$ to $79\%$, whereas with EWC catastrophic forgetting was mitigated and accuracy was $82\%$.
In all experiments with EWC, the original task accuracy was significantly higher than fine-tuning without regularization and fine-tuning with L2 regularization. These gains were attained while maintaining similar performance on the fine-tuning data.
Plotting the Pareto frontier, we show that equivalent gains in accuracy can be made with less forgetting of the original dataset.
Furthermore, we demonstrate that fine-tuning methods can be combined with POE and DFL, yielding improvements on both the original task as well as fine-tuning data used for debiasing.
Similar patterns were observed when fine-tuning MultiNLI models with lexical bias evaluation datasets \citep{Naik2018}.

\section{Mitigating biases with fine-tuning}

Fine-tuning broadly refers to approaches where a model is initially trained on one dataset and then further improved by training on another. We refer to these datasets as fine-tuning training and test data as \emph{FT-train} and \emph{FT-test} respectively.
This technique is commonly used to mitigate model biases \cite{Liu2019InoculationDatasets}, where the original data, while useful in model training, often contain biases, which are addressed by further training the model on a small set of instances targeting these biases.
%{
%\color{purple} 
%Fine-tuning requires no change to the underlying model architecture; whereas in comparison, adversarial training techniques require additional components to model the bias \citep{Mahabadi2020End-to-EndCorpora}.
%While the hypothesis-only bias can expressed and mitigated through both techniques, not all biases or model limitations can be readily isolated through architectural changes alone. 
%When fine-tuning, the behavior to be mitigated is expressed in a targeted set of data that is used adapt the model parameters allowing limitations and biases to expressed without a high engineering cost.
%}
Fine-tuning to mitigate bias, however, can result in model parameters over-adjusting to the instances targeting it, reducing the accuracy on the original task, referred to as catastrophic forgetting \cite{French1999}. 
To ameliorate this issue, one can regularize the parameter updates so that they do not deviate too much from the original training, similar to the intuition behind multi-task training approaches \citep{Ruder2017AnNetworks}.

%To mitigate this issue, a common approach in the literature\todo{would be good to give pointers here, I don't have any OTOh though. Or rephrase to avoid}

%Limitations in the data for the first task cause some parameters that enable general-domain behaviours to be dominated by those that replicate phenomena captured in the training data.
%Further training of the model by fine-tuning on data from a different distribution regularizes model parameters through exposure to wider range of phenomena.
% certain biases.
%the sentence pair interaction between the claim and the evidence would be weak. 
%While these parameters would not be critical for the biased test set used to compute the EWC penalty, they would be updated during fine-tuning, allowing fine-tuning on the symmetric data subset to debias the model with minimal change to the model's behavior on the original FEVER data.

Elastic Weight Consolidation \cite[EWC]{Kirkpatrick2017} penalizes parameter updates according to the model's sensitivity to changes in these parameters. % instances sampled from the original dataset.
The model sensitivity is estimated by the Fisher information matrix, which describes the model's expected sensitivity to a change in parameters, and near the (local) minimum of the loss function used for training is equivalent to the second-order derivative $F = \mathbb{E}_{(x,y)\sim\mathcal{D}_{original}}\lbrack \nabla^2 \log p(y|x;\theta) \rbrack$. 
When fine-tuning with EWC (which we refer to as FT+EWC), the Fisher information is used to elastically scale the cost of updating parameters $\theta_i$ from the original value $\theta_i^*$, controlled by the $\lambda$ hyper-parameter, as follows: 
\begin{equation}
\mathcal{L}(\theta) = \mathcal{L}_{FT}(\theta) + \sum_i {\frac{\lambda}{2}F_{i,i}(\theta_i - \theta_{i}^{*})^2}
\label{eq:ewc}
\end{equation}
For efficiency, we use the \emph{empirical Fisher} \citep{Martens2014NewMethod}:  diagonal elements are approximated through squaring first-order gradients from a sample of instances, recomputed before each epoch.
If the Fisher information is not used (i.e.\ $F_{i,i}=1$), Eq.~\ref{eq:ewc} is equivalent to L2 regularization (which we refer to as FT+L2). 

%This constraint would allow models to be fine-tuned on data from a different domain, minimizing changes to parameters for the original task.
%This is especially pertinent as 
%While these parameters would not be critical for the biased test set used to compute the EWC penalty, they would be updated during fine-tuning, allowing fine-tuning on the symmetric data subset to debias the model with minimal change to the model's behavior on the original FEVER data.

%\citet{Thompson2019} and \citet{Saunders2019DomainTranslation} overcome catastrophic forgetting when fine-tuning Neural Machine Translation models (NMT) undergoing domain adaptation through applying an additional loss term that penalizes changing parameters required for the original task. 

% what is the fisher information - and what's the intuition of it. why does it matter
% justify application of this techniq
% connection to multitask learning and continuous learning
% Issues with scaling to more tasks in fine-tuning problems in existing work 

%We justify application of this technique, is that a model trained on biased data will learn most parameters appropriately for natural language understanding - however parameters that capture the sentence pair interaction between the claim and the evidence would be weak. 
%While these parameters would not be critical for the biased test set used to compute the EWC penalty, they would be updated during fine-tuning, allowing fine-tuning on the symmetric data subset to debias the model with minimal change to the model's behavior on the original FEVER data.

\section{Experimental setup}
\label{sec:setup}

We assess the application of EWC to minimize catastrophic forgetting when mitigating model biases in the context of two sentence-pair classification tasks: fact verification and natural language inference. 
We compare the untreated model (original), fine-tuning (FT), FT+EWC, FT+L2, and merging instances from the FT-train dataset when training (Merged).
Each model is first trained using the original dataset and splits from the respective task, using the AllenNLP implementations \citep{Gardner2017} with default hyper-parameters and tokenize with SpaCy or pre-trained transformer tokenizers.
We train five random initializations of each model, reporting the mean accuracy, standard deviation, and $p$-value with an unpaired $t$-test.
For fine-tuning, the learning rate, regularization strength $\lambda$, and number of epochs are selected through 5-fold cross-validation on the FT-train data, selecting the model with the highest FT-train accuracy. 
30 hyper parameter choices were evaluated with grid search over 10 choices for regularization strength between $10^6$ and $10^8$ and 3 choices of learning rates in $\{2\cdot 10^{-6},4 \cdot 10^{-6}, 6\cdot 10^{-6} \}$ for transformer models and $\{2\cdot10^{-4}, 4\cdot10^{-4},6\cdot10^{-4}\}$ for ESIM models. We trained the models for a max of 8 epochs.  For the transformer-based models, the highest cross validation accuracy on the  FT-train dataset was achieved with $LR=4\cdot 10^{-6}$, $\lambda=10^7$ and 6 epochs. For the ESIM-based models, the highest FT-train accuracy was achieved with $LR=2\cdot 10^{-4}$, $\lambda=10^7$ and 5 epochs. 
Full hyper-parameter choices are in Appendix~\ref{app:hyper}.

\paragraph{Mitigating hypothesis only bias in fact verification:} 
The FEVER\footnote{\url{https://fever.ai/}} task \citep{Thorne2018a} is to predict whether short factoid sentences called claims are Supported or Refuted against evidence (in the form of sentences from Wikipedia) or whether there is not enough info (NEI).
When training the models, the NEI instances by definition don't have evidence: we sample negative instances for these with random sentences from the Wikipedia page closest to the claim using TF$\cdot$IDF. 
This preprocessing is the same as \citet{Thorne2018a}.
\citet{Schuster2019} identified a bias where the label for some claims can be predicted without the need for evidence. 
To evaluate this bias, they released\footnote{\url{https://github.com/TalSchuster/FeverSymmetric/}} a set of 1420 symmetric counterfactual instances where each claim is supported by one Wikipedia passage and refuted by another (approximately $1\%$ of the FEVER dataset). 
This is mitigating the claim-only bias by reducing the mutual information between claims and labels.
The availability of counterfactual data meant that it is possible to experiment with fine-tuning as a mitigation strategy, using the published dev and test data as FT-train and FT-test respectively.
Following \citet{Schuster2019}'s evaluation, we train two ESIM (Enhanced LSTM) variants \citep{Chen2016}, and a BERT \citep{Devlin2019} transformer. 
We also evaluate RoBERTa \citep{Liu2019RobustlyOptimized}, as it has been shown to be more robust to adversarial testing \citep{Bartolo2020BeatComprehension}.

\paragraph{Mitigating model limitations in NLI stress tests:}
The MultiNLI\footnote{\url{https://cims.nyu.edu/~sbowman/multinli/}} task \citep{Williams2018} requires systems to predict whether a hypothesis is entailed by a premise.
\citet{Naik2018} identify limitations of models trained on this dataset where 6 phenomena such as \emph{lexical overlap}, \emph{numerical reasoning} and presence of \emph{antonyms} were evaluated with `stress-tests'. In this paper, we report on \emph{antonyms} and \emph{numerical reasoning} as these stress-tests exhibited catastrophic forgetting when used to fine-tune models \citep{Liu2019InoculationDatasets}. To this end, we do not evaluate on HANS \cite{ThomasMcCoy2020}, as high accuracies can be attained without forgetting.

Like \citet{Liu2019InoculationDatasets}, we mitigate these biases through fine-tuning both an ESIM \citep{Chen2016} model on stress-test data\footnote{\url{https://abhilasharavichander.github.io/NLI_StressTest/}}.
Each contains a small number of procedurally generated instances (between 1500-9800) that specifically target one of these phenomena. 
We evaluate FT and FT+EWC using the same methodology, controlling the number of instances, sampled at random, in FT-train and report the change in accuracy on the FT-test and original test sets.  

\section{Results}% main results table goes here (force on page 3)
\input{table_finetuning_withconcat}

\subsection{Fact verification}

%Using EWC minimised catastrophic forgetting while achieving higher gains in accuracy on the symmetric dataset than instance re-weighting. \todo[color=green]{JT: rephrase this after computing curve}
%still maintaining competitive accuracy on the symmetric dataset.\todo[color=green]{JT: check. Need to compare even if loosely to Schuster/Liu et al fine tuning methods} 
%Prior to fine-tuning, 
%We report the 3-class model accuracy on both the FEVER and symmetric datasets in %the `baseline' column of Table~\ref{tab:test}.

%All fine-tuned models improved accuracy on the symmetric dataset at the expense of forgetting the original FEVER task.

Fine-tuning the models, rather than merging datasets, yielded the greatest improvements in accuracy on FT-test. All improvements from the untreated model were significant ($p<0.05$, denoted $\#$).  Without L2 or EWC, catastrophic forgetting occurs due to the shift in label distribution between the FEVER and FT-train dataset, which only contains 2 of the original 3 label classes.% (see Section~\ref{sec:setup}).

Both L2 and EWC reduced catastrophic forgetting. Improvements on the original task are significant (denoted $*$) compared to FT. However, EWC regularization retained more of the original task accuracy than L2 for all models; this was also significant (denoted $\dagger$). 
In all cases, there is a trade-off between original and fine-tuning task accuracies. With regularization, the FT accuracy was higher than FT+L2 and FT+EWC (with the exception of ESIM+ELMo). However, the deterioration from FT when regularizing was not significant ($p>0.05$, denoted $\diamondsuit$). 
Furthermore, for the highest performing model (RoBERTa base), the deterioration of using FT+EWC against FT+L2 was also not significant (denoted $\heartsuit$). 

%For the FT+EWC models, the accuracy on the FEVER test set is significantly higher than FT $(p<0.01)$ showing that catastrophic forgetting is reduced when using elastic weight consolidation. 
%While FT-test accuracy for FT+EWC was not as high as FT, the differences were not significant (except for BERT) and all scores for the FT-test for FT+EWC remain significantly higher than for the original, untreated model ($p<0.01$).
%
%With the exception of the ELMo model, %where the accuracy on the symmetric dataset increased at the expense of the FEVER task accuracy, 
%Incorporating an L2 penalty (FT+L2) had no significant harm $(p>0.05)$ on the FT-test accuracy compared to FT. However, L2 regularization yielded no significant improvement to catastrophic forgetting on the original FEVER task compared to FT.

Training a model where the FEVER training and FT-train were merged yielded modest improvements on the FT-test without harming the original FEVER task accuracy. 
We attribute this to the impact of these 700 instances being diluted by the large number of training instances in FEVER (FT-train is $<$1\% the size of FEVER). 
%For the RoBERTa model, accuracy on FT-test increases to $87.03\%$. While this does not exceed the accuracy of any of the fine-tuned models, it is not significantly worse than FT+EWC ($p=0.06$).

% where all decreases in accuracy were significant $(p<0.01)$.

\begin{figure}[t]
% \subfloat[Vanilla models]{
% \includegraphics[width=\linewidth]{graphs/pareto_new.pdf}
% %
% \label{fig:pareto}
% }
% \vfil
% \subfloat[Combined with bias mitigation]{
\includegraphics[width=\linewidth]{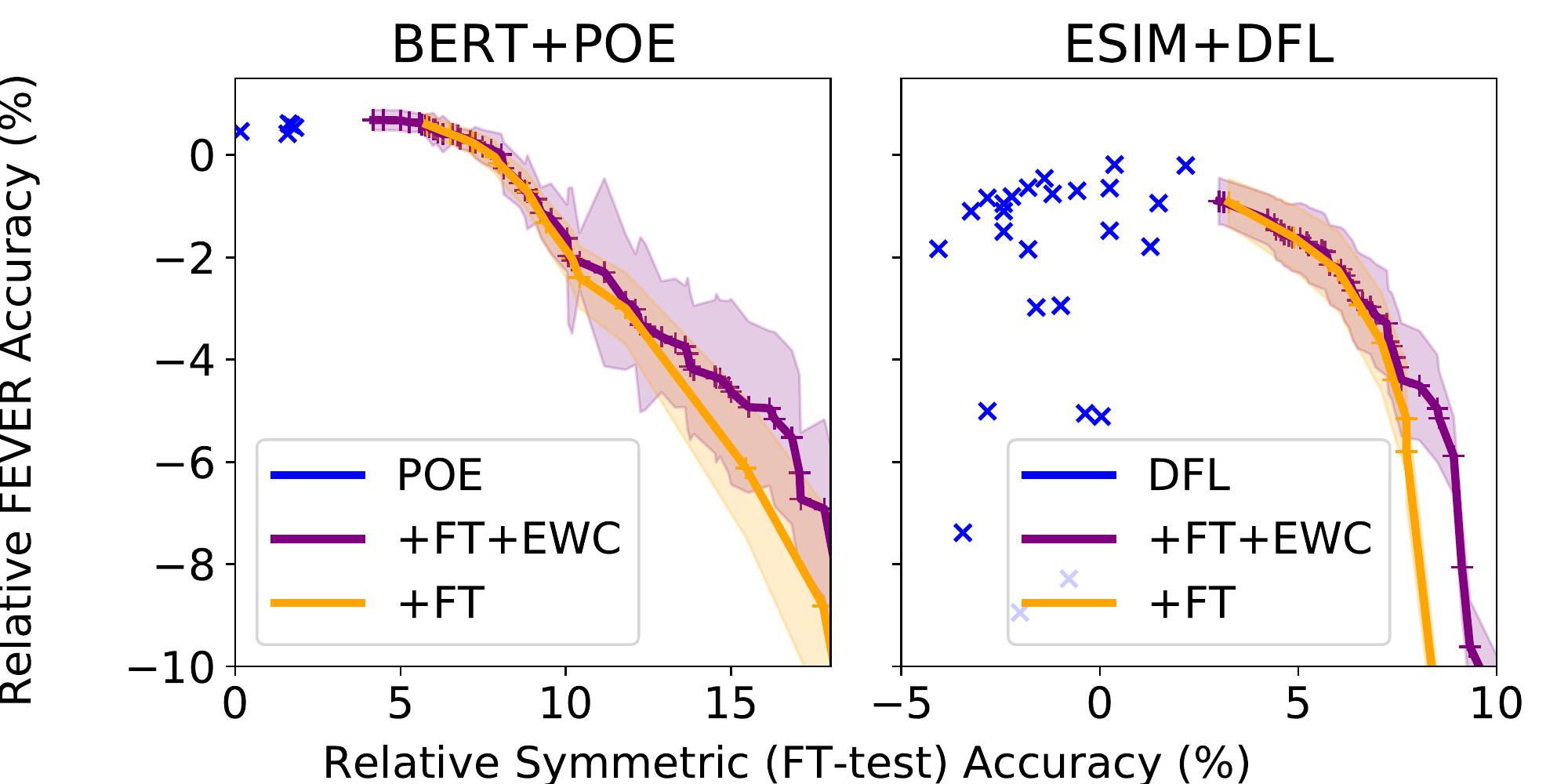}
%\caption{Pareto frontier fine-tuning a BERT model debiased with POE and an ESIM debiased with DFL. Accuracy is relative to models with no debiasing. \label{fig:pareto2}}

%}
\caption{Pareto frontiers of fine-tuning BERT and ESIM models showing FT+EWC dominates FT. }
\label{fig:pareto}
\end{figure}

%While a minor reduction in accuracy is observed on the symmetric test set compared to fine-tuning without regularization for the ESIM+GloVe, for BERT and RoBERTa it is not significant $(p>0.05)$. For the ESIM+ELMo model, there was no reduction in performance. %Even with this minor reduction, all accuracies reported on the symmetric test after fine-tuning with EWC were significantly higher than applying no treatment.

\subsection{Combining FT and bias modeling}
Fine-tuning can be applied to any task using a small amount of bias-mitigating labeled data, whereas explicit modeling of hypothesis-only biases \cite{Mahabadi2020End-to-EndCorpora} requires architectural changes that are specific to the task and model. 
%\cite{Mahabadi2020End-to-EndCorpora}.
%To mitigate hypothesis-only bias during training, the encoded hypothesis is used to predict a label and this is used to weight the loss for instances according to the bias. 
We consider two techniques from \citet{Mahabadi2020End-to-EndCorpora}: Product of Experts (PoE), which multiplies the sentence-pair label probabilities with hypothesis-only label probabilities, and Debiased Focal Loss (DFL)
which explicitly modulates the loss of instances according to the accuracy of a hypothesis-only classifier. 
For both ESIM and BERT,  accuracy, when trained with PoE, was stable across different choices of hyper-parameters,
%resulted in similar accuracy, 
whereas some hyper-parameter choices for DFL resulted in lower accuracy on both tasks.
We report results for BERT+PoE and ESIM+DFL as these were best.
%; for readability,  the figure only shows ESIM+DFL results that were in the same range as ones obtained with finetuning, omitting many of the worse results obtained.

The trade-off between accuracy on the original and FT-test datasets is visualized in Figure~\ref{fig:pareto} indicating the change in accuracy for both bias modeling techniques in isolation, as well as in combination with fine-tuning.\footnote{All Pareto frontier lines are the result of sweeping the following: for POE, $\beta$; for DFL, $\beta$ and $\gamma$; for FT, learning rate; and for FT+EWC, learning rate + EWC strength ($\lambda$ in Equation~\ref{eq:ewc}) in Appendix~\ref{app:sweep}.}
This further shows that FT+EWC Pareto dominates FT for both the ESIM and BERT model. With EWC, equivalent gains on the symmetric FT-test can be attained with a lower reduction in accuracy on the original FEVER task.

%indicate the same trade offs as models without unsupervised mitigation. Where model biases can be encoded through model architecture choices, fine-tuning can be used to control the trade-off in accuracy between the original and FT tasks. 

\subsection{Natural Language Inference (NLI)}

%Without regularization, we found that the procedural generation of these instances induces different patterns and biases that models exploited the cost of the forgetting the original task if fine-tuned without regularization. %todo{on a first pass this sentence sounds morte about the data than the experiments, might be good to rephrase it} 
%In our experiments incorporating EWC (see Figure~\ref{fig:nli}), the models are `inoculated' against the patterns in the stress-test data -- improving accuracies for the two challenges from $33\%$ for both models to above $70\%$ -- while catastrophic forgetting is minimized.
%EWC was most effective for the `antonym' challenge which had the largest domain shift from the original data due to all instances being labelled as contradiction as without regularization, the model just learned to predict the majority label.

\begin{figure}[]
\includegraphics[width=\linewidth]{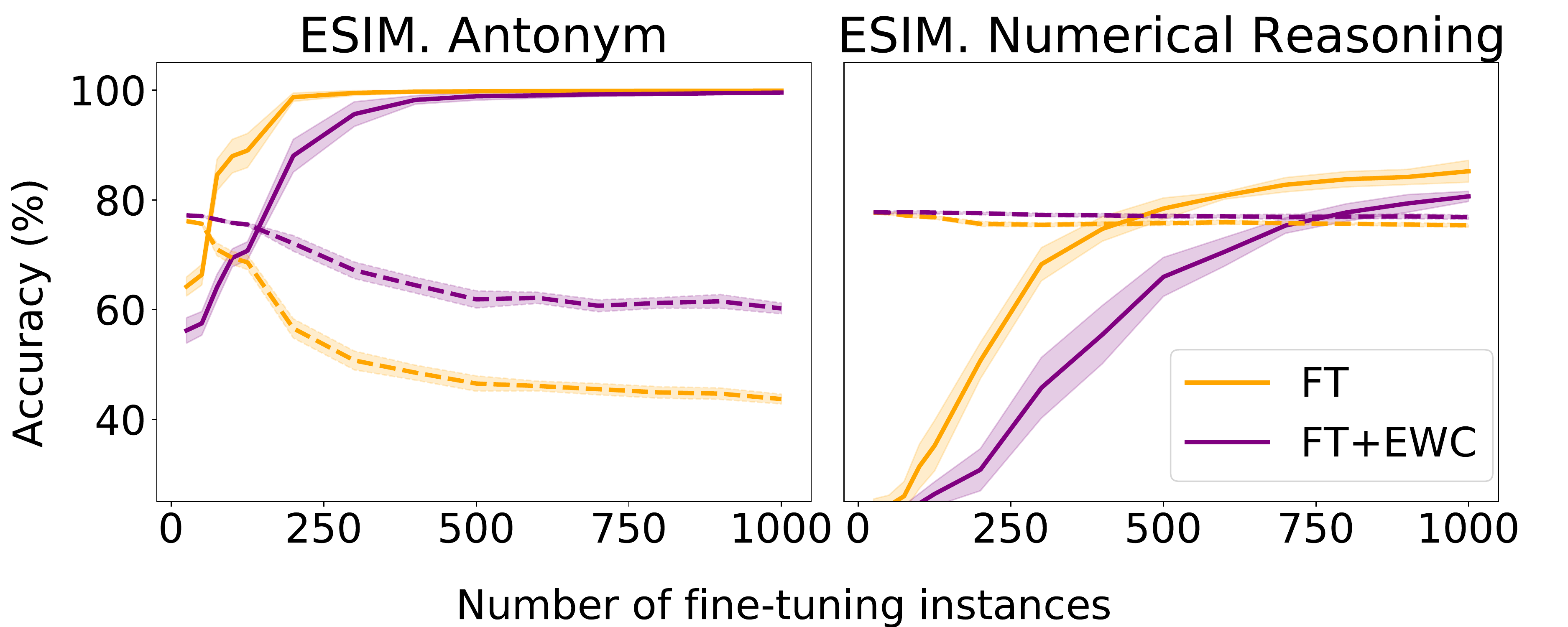}
\caption{Training curves fine tuning MultiNLI with stress-test data. Solid lines indicate challenge dataset accuracy. Dashed lines indicate MultiNLI accuracy. \label{fig:nli}}
\end{figure}

In a separate experiment, we apply EWC to a different domain. We inoculate biases on the ESIM model for natural language inference reported in \citet{Liu2019InoculationDatasets}. For both \emph{Antonym} and \emph{Numerical Reasoning} challenges, MultiNLI accuracy was higher with FT+EWC compared to FT (dashed lines in Figure~\ref{fig:nli}). 

\paragraph{Antonym challenge:} The ESIM model was sensitive to fine-tuning, attaining near perfect accuracy (top row of Figure~\ref{fig:nli}) on the FT-test data. 
The antonym stress-test only contains instances labeled \emph{contradiction}: a change in label distribution that causes catastrophic forgetting.
%(this was also observed by \citet{Liu2019InoculationDatasets}).
Without EWC, accuracy on MultiNLI fell to just above chance levels as the model learned only to predict contradiction (yellow dashed line).
%Choosing an appropriate EWC penalty controlled catastrophic forgetting: a low 
However, using an appropriate
EWC penalty attained near-perfect accuracy with a smaller reduction in MultiNLI accuracy (purple dashed line). 
%and with a high EWC penalty, accuracy was still $>93\%$ for both DA and ESIM with lower reduction in accuracy for the MultiNLI task (yellow dashed lines).

%With a low EWC penalty, both models attained $>99\%$ accuracy on the challenge data, but MultiNLI accuracy only fell to $54.4\%$ for the DA and $60.2\%$ for the ESIM. With a higher EWC penalty the stress-test accuracy was $93.0\%$ for the DA and $98.5\%$ for the ESIM. Accuracy on MultiNLI fell to $66.2\%$ and $68.6\%$ for the two models respectively. 
%While \citet{Liu2019InoculationDatasets} do not show empirical evaluation for this challenge; their paper remarks that fine-tuning on data with a limited number of patterns is not informative due to the models learning to predict the majority label. This is avoided with EWC.

\paragraph{Numerical reasoning challenge:} The ESIM model was sensitive to fine-tuning to introduce numerical reasoning behaviours to the model. As the difference in label distribution in the inoculation dataset was less severe than the Antonym dataset, the catastrophic forgetting was less severe.
Nevertheless, FT+EWC minimized catastrophic forgetting at the expense of reducing sample efficiency: accuracy on MultiNLI fell from $77.9\%$ to $75.4\%$ without EWC and $76.8\%$ with EWC.

\section{Conclusions}
Fine-tuning can be used to mitigate model bias but has the risk that
%Fine-tuning models is a multi-objective optimization problem, trading-off accuracy on one dataset for accuracy on another,  as 
the model catastrophically forgets the data it was originally trained on.
%We evaluated fine-tuning sentence-pair classifiers to mitigate biases and model limitations. 
%For sentence-pair classifiers, fine-tuning models allows new behaviours such as numerical reasoning to be learned from stress-test data \citep{Liu2019InoculationDatasets}. 
%In this paper, we show that, in the context of fact verification, fine-tuning mitigates hypothesis-only bias \citep{Gururangan2018,Poliak2018}.
Incorporating elastic weight consolidation (EWC) when fine-tuning minimizes catastrophic forgetting, yielding higher accuracy on the original task.
We show this holds for both the NLI stress-tests, as well %as a novel evaluation,
debiasing fact-verification systems \citep{Schuster2019} through fine-tuning.
%Furthermore, we demonstrate complementary gains between EWC and unsupervised bias mitigation techniques \citep{Mahabadi2020End-to-EndCorpora}.

%While previous work by \cite{Liu2019InoculationDatasets} show that fine-tuning models is one way to remove bias, in some cases where the data distributions differ, this removal of bias comes at the expense of catastrophic forgetting.
%We showed that an elastic weight consolidation penalty is sufficient to minimize catastrophic forgetting when using fine-tuning to mitigate biases in sentence pair classification tasks{\color{purple}, even when combined with other adversarial training techniques.}

\section*{Acknowledgements}

James Thorne is funded by an Amazon Alexa Graduate Fellowship. Andreas Vlachos is supported by the ERC grant AVeriTeC. Thank you to Danielle Saunders for the insightful discussions.

\bibliography{references}
\bibliographystyle{acl_natbib}

\input{appendix}

%The availability of counterfactual data meant that it was possible to experiment with fine-tuning as a mitigation strategy. 
%However, this type of debiased data is not available for other sentence-pair fact verification tasks. 
%While LiarPlus \citep{Alhindi2018} and MultiFC \citep{Augenstein2019} both have a similar task signature with claims and additional information provided as text, the text is not evidential (i.e. not containing a premise that would support or refute the claim).
%Instead, the additional text in these datasets serves as explanation and related information. 
%This is indicated by experimental results in Table~\ref{tab:bias} where classification models have higher accuracy than a text-pair classification when only using the claim text.
%Future work will consider how biases in these datasets could be mitigated, dependent on the availability of counterfactual data.

%\section{Hyper-parameters}
%\label{app:hyper}

\end{document}

%% file: table_finetuning_withconcat.tex
\begin{table*}[]
\centering
\begin{tabular}{lccccc}
\toprule
\multicolumn{1}{c}{\multirow{2}{*}{\textbf{Model}}}  & \multicolumn{5}{c}{\textbf{FEVER Dataset (Original Task) Accuracy (\%)}}                                 \\ 
 & \textbf{Original} & \textbf{Merged} & \textbf{FineTune} & \textbf{FineTune+L2} & \textbf{FineTune+EWC} \\ 
\midrule
ESIM+GloVe             & $79.94 \pm 0.4$ & $79.57 \pm 0.4$ & $70.78 \pm 1.1$  & $73.29 \pm 0.4^*$ & $74.64 \pm 0.7^{*\dagger}$  \\
ESIM+ELMo              & $80.15 \pm 0.2$ & $80.33 \pm 0.8$ & $76.45 \pm 0.8$  & $73.72 \pm 0.6^*$ & $78.09 \pm 0.4^{*\dagger}$  \\
BERT Base              & $86.88 \pm 0.5$ & $86.87 \pm 0.5$ & $78.82 \pm 0.9$  & $79.90 \pm 1.4^*$ & $82.23 \pm 1.1^{*\dagger}$  \\
RoBERTa Base           & $88.12 \pm 0.3$ & $88.11 \pm 0.1$ & $82.51 \pm 1.5$  & $83.14 \pm 1.4^*$ & $85.12 \pm 1.1^{*\dagger}$  \\ \midrule
                       & \multicolumn{5}{c}{\textbf{Symmetric Dataset (Fine-tuning Task) Accuracy (\%)}}                         \\ \midrule
ESIM+GloVe             & $68.37 \pm 1.0$ & $69.35 \pm 0.5$ & $74.21 \pm 1.3^{\#}$ & $73.34 \pm 1.2^{\#\diamondsuit}$ & $73.20 \pm 1.4^{\#\diamondsuit\heartsuit}$  \\
ESIM+ELMo              & $64.04 \pm 0.7$ & $66.46 \pm 1.3^{\#}$ & $68.68 \pm 0.7^{\#}$ & $70.31 \pm 0.5^{\#}$ & $69.16 \pm 0.7^{\#}$  \\
BERT Base              & $74.77 \pm 1.4$ & $79.24 \pm 0.7^{\#}$ & $87.07 \pm 0.6^{\#}$ & $86.66 \pm 0.4^{\#\diamondsuit}$ & $85.11 \pm 0.4^{\#}$  \\
RoBERTa Base           & $78.34 \pm 0.2$ & $87.03 \pm 2.3^{\#}$ & $91.01 \pm 0.6^{\#}$ & $90.98 \pm 0.5^{\#\diamondsuit}$ & $89.63 \pm 1.3^{\#\diamondsuit\heartsuit}$  \\ \bottomrule
\end{tabular}

\caption{\label{tab:test} Bias mitigation for FEVER classifiers comparing no treatment (original), against merging from instances from the FT-train with the original task training dataset (Merged) and FineTuning (with EWC and L2). Improvements $p<0.05$ are marked with the following symbols: $*$ against FT, $\dagger$ against FT+L2, $\#$ against original. Deteriorations $p>0.05$ on the symmetric dataset are marked with $\diamondsuit$ against FT and $\heartsuit$ against FT+L2}

\end{table*}

%% file: appendix.tex
\appendix 
\section{Hypothesis-only bias in other fact-verification datasets}
\label{app:datasets}
We use the FEVER dataset in this paper due to the availability of the symmetric counterfactual data released by \citet{Schuster2019}. 
Other datasets for fake news detection and fact verification also have the same task signature (sentence-pair classification with a claim and another sentence), but second sentence is not evidential.
All tasks exhibit a hypothesis-only bias where information from the claim can be used to predict the label without considering the second sentence in the sentence-pair.
If there was no mutual information between claims (without evidence) and labels, this should be $33\%$.

For FEVER, this bias is introduced through synthetic generation of the claims and is more problematic than the biases that occur in the datasets consisting of naturally occurring claims. 
In Liar and MultiFC, the claims arise from real-world events and the biases in the data reflect political viewpoints, rather than cognitive shortcuts taken by the FEVER annotators.

The RoBERTa model, trained on only the claims outperforms the sentence-pair setup for both the MultiFC \cite{Augenstein2019} and Liar-Plus tasks \cite{Alhindi2018} whereas for the the FEVER data the sentence-pair accuracy is higher.
As FEVER is the only task that requires the use of evidence for classification, this is expected.

\begin{table}[h!]

\centering
\begin{tabular}{lcc}
\toprule
\multicolumn{1}{c}{\multirow{2}{*}{\textbf{Dataset}}} & \multicolumn{2}{c}{\textbf{Accuracy (\%)}}    \\
\multicolumn{1}{c}{}                                  & \textbf{Claim Only} & \textbf{Sentence Pair} \\ \toprule
Liar-Plus                                             & 28.74               & 20.48                  \\
Liar-Plus (binary)                                    & 72.59               & 70.48    \\

MultiFC &  46.02                  & 44.83                  \\\midrule
FEVER & 61.50 & 88.93  \\
FEVER (2-way) & 79.09 & 92.24 \\
\bottomrule
\end{tabular}
\caption{Validation accuracy for claim-only vs sentence pair classification for fact verification datasets trained on RoBERTa\label{tab:bias}. For 2-way fever we discard instances labelled \textsc{NotEnoughInfo}. For binary Liar-Plus, we map all positive labels to true and all negative labels to false and discard neutral instances.}

\end{table}

\section{Compute infrastructure}
All experiments were performed on a single workstation with a single Xeon E5-2630 CPU, 64GB RAM and an Nvidia 1080Ti GPU.

\section{Average run-time for fine-tuning}
% Estimating the Fisher matrix diagonal takes approximately 0.2 seconds for the DA model using 2000 instances sampled from the original dataset. Average training duration (excluding estimating Fisher information) was 3.5 seconds in total with 700 instances.

Estimating the Fisher matrix diagonal takes approximately 2 seconds for the ESIM model using 2000 instances sampled from the original dataset. Average training duration (excluding estimating Fisher information) was 11 seconds in total with 700 instances.

Estimating the Fisher matrix diagonal takes approximately 25 seconds for the BERT and RoBERTa models using 2000 instances sampled from the original dataset. Average training duration (excluding estimating Fisher information) was 2 minutes 20 seconds in total with 700 instances.

\section{Hyper-parameter configurations}
\label{app:hyper}
\subsection{Base models}
For the base-models, the default hyperparemters in AllenNLP are used for the ESIM, BERT and RoBERTa models.

\paragraph{ESIM}
\begin{itemize}[noitemsep,nolistsep]
    \item Embedding dimension: 300, bidirectional
    \item Dropout: 0.5
    \item Optimizer: Adam
    \item Gradient Norm: 10.0
    \item Batch Size: 64
    \item Learning Rate: 0.0004
    \item Learning Rate Schedule: reduce on plateau, patience 0, factor 0.5
    \item Number of Epochs: 75
    \item Early Stopping: Patience 10
\end{itemize}

\paragraph{BERT+RoBERTa}
\begin{itemize}[noitemsep,nolistsep]
    \item Embedding dimension: 768
    \item Optimizer: AdamW
    \item Gradient Norm: 10.0
    \item Batch Size: 8
    \item Learning Rate: 0.0004
    \item Learning Rate Schedule: slanted triangular, cut frac 0.06
    \item Number of Epochs: 5
    \item Early Stopping: Patience 0
\end{itemize}

% For DA and ESIM, embedding dimension is 300 for ESIM and 200 fof, gradient norm is 10.0, and ADAM learning rate is 0.0004. Models are trained for 75 epochs with a patience of 10 epochs

\subsection{Fine-tuning without EWC}
\paragraph{ESIM}
\begin{itemize}[noitemsep,nolistsep]
    \item FT Learning Rate: 0.0002
    \item FT Epochs: 8
\end{itemize}

\paragraph{BERT}
\begin{itemize}[noitemsep,nolistsep]
    \item FT Learning Rate: 0.000004
    \item FT Epochs: 6
\end{itemize}

\paragraph{RoBERTa}
\begin{itemize}[noitemsep,nolistsep]
    \item FT Learning Rate: 0.000004
    \item FT Epochs: 7
\end{itemize}

\subsection{Fine-tuning using EWC}
\paragraph{ESIM}
\begin{itemize}[noitemsep,nolistsep]
    \item FT Learning Rate: 0.0002
    \item FT Epochs: 5
    \item EWC: 10000000
\end{itemize}

\paragraph{BERT+RoBERTa}
\begin{itemize}[noitemsep,nolistsep]
    \item FT Learning Rate: 0.000004
    \item FT Epochs: 6
    \item EWC: 10000000
\end{itemize}

\subsection{Unsupervised bias mitigation}
Hyperparameters: $\beta$ controls the weight update for the hypothesis-only model and $\gamma$ controls the modulation of hypothesis-only model in the loss function.
\begin{itemize}[noitemsep,nolistsep]
    \item POE (BERT) $\beta=0.4$
    \item POE (ESIM) $\beta=0.05$
    \item DFL (BERT) $\beta=0.4, \gamma=0.6$
    \item DFL (ESIM) $\beta=0.05, \gamma=2.0$
\end{itemize}

\subsection{Search bounds for fine-tuning}
\label{app:hyper_ft}
Approximately 30 configurations were considered (Cartesian product of LR+EWC). The best performing system was selected through max accuracy on the FT cross validation dataset through 5 fold cross validation. 

\begin{itemize}
    \item EWC $\{ 10^{6},2\cdot10^{6}, 4\cdot10^{6}, 8\cdot10^{6}, 10^7, 2\cdot10^7$,
    $4\cdot10^7, 6\cdot10^7, 8\cdot10^7, 10^8 \}$
    \item Fine-tuning learning rate (ESIMs): $\{0.0002, 0.0004, 0.0006\}$
    \item Fine-tuning learning rate (Transformer): $\{ 0.000002, 0.000004, 0.000006\}$
    \item Epochs: Up to 10 epochs cross valudating on the FT-training dataset
\end{itemize}

\subsection{Search bounds for end2end bias mitigation}
\label{app:hyper_e2e}
We used the same range of values published by \citet{Mahabadi2020End-to-EndCorpora}. For DFL, we performed a grid search of every value totalling 30 trials.

$\gamma \in \{0.02, 0.05, 0.1, 0.6, 2.0, 4.0, 5.0\}$

$\beta \in \{0.05,0.2,0.4,0.8,1.0\}$

% \section{Dataset Statistics}
% \subsection{FEVER}
% We used the published FEVER dataset partitions which are disjoint by entity.

% \begin{itemize}
%     \item Train: 142k
%     \item Dev: 10k
%     \item Test: 10k
% \end{itemize}

% \subsection{Symmetric}
% We used the published symmetric dataset partitions which are disjoint by entity.

% \begin{itemize}
%     \item Dev: 700 (used for training + cross validation)
%     \item Test: 700 (used for testing)
% \end{itemize}

% \subsection{MultiNLI}
% We used the published FEVER dataset partitions which are disjoint by entity.

% \begin{itemize}
%     \item Train: 393k
%     \item Dev: 10k
%     \item Test: 10k
% \end{itemize}

\subsection{Stress test sizes}
Following the evaluation of \cite{Liu2019InoculationDatasets} we vary the number of instances from the stress test (between 1500-9800) instances.
To plot Figure~\ref{fig:nli}, we use 25, 50, 75, 100, 250, 400, 500, 600, 700, 800, 900, 1000 instances in our evaluation.

\subsection{Pareto frontier sweeps}
\label{app:sweep}
The Pareto frontier sweeps in Figure~\ref{fig:pareto} were generated by plotting all hyperparameters from Appendix~\ref{app:hyper_ft}. The end2end (blue) crosses are plotted by plotting the choices from Appendix~\ref{app:hyper_e2e} in combination with the best FT model.